\documentclass[preprint]{article} 

\PassOptionsToPackage{numbers}{natbib}
\usepackage{preprint_aug2025}

\usepackage[utf8]{inputenc} 
\usepackage[T1]{fontenc}    
\usepackage{hyperref}       
\usepackage{url}            
\usepackage{booktabs}       
\usepackage{amsfonts}       
\usepackage{nicefrac}       
\usepackage{microtype}      
\usepackage{xcolor}         
\usepackage[pdftex]{graphicx}

\title{LLM Assertiveness can be Mechanistically Decomposed into Emotional and Logical Components}

\author{
  Hikaru Tsujimura. \thanks{Contribution: Lead on Conceptualization, Methodology, Investigation, and Writing – Original Draft. 
} \\
  Department of Psychology, Cardiff University, Cardiff, UK \\
  \texttt{htsujimu@gmail.com} \\
  \And
  Arush Tagade. \thanks{Contribution: Writing – Review \& Editing (equal); critical internal review and feedback.} \\
  Department of Computer Science, George Washington University, Washington, DC, USA\\
  \texttt{arush.tagade@gwu.edu} \\
}

\begin{document}

\maketitle

\begin{abstract}

Large Language Models (LLMs) often display overconfidence, presenting information with unwarranted certainty in high-stakes contexts. We investigate the internal basis of this behavior via mechanistic interpretability. Using open-sourced Llama 3.2 models fine-tuned on human annotated assertiveness datasets, we extract residual activations across all layers, and compute similarity metrics to localize assertive representations. Our analysis identifies layers most sensitive to assertiveness contrasts and reveals that high-assertive representations decompose into two orthogonal sub-components of emotional and logical clusters-paralleling the dual-route Elaboration Likelihood Model in Psychology. Steering vectors derived from these sub-components show distinct causal effects: emotional vectors broadly influence prediction accuracy, while logical vectors exert more localized effects. These findings provide mechanistic evidence for the multi-component structure of LLM assertiveness and highlight avenues for mitigating overconfident behavior.

\end{abstract}

\section{Introduction}

Large Language Models (LLMs) are increasingly deployed in high-stakes domains such as law \citep{lai_large_2024}, politics \citep{bai_llm-generated_2025}, education \citep{trust_editorial_nodate}, and public health \citep{sharma_cognitive_2023}. A major concern of LLM usage is that they often produce overconfident statements, overstating information regardless of its factual basis \citep{ghafouri_epistemic_nodate}. Such behavior can mislead users, amplify bias, and result in poor decisions with serious societal consequences.

Prior work \citep{leng_taming_2025, zhou_steerconf_2025, wen_mitigating_nodate} has quantified LLM overconfidence through linguistic assertiveness cues such as “highly certain,” or “I am confident that”. While prominent in human communication, it remains unclear how these cues are internally represented, and whether LLMs treat assertiveness as a single construct or as multiple, separable components.

Beyond sensitivity to linguistic cues, assertiveness itself plays mixed roles. In human studies, some associate it with certainty and confidence \citep{windschitl_measuring_1996, karelitz_you_2004}, while others link it to persuasion \citep{paradeda_role_2019}. Distinguishing these roles is critical: if models primarily respond to one, or encode them separately, this implies distinct internal mechanisms.

Mechanistic interpretability provides a way to test this (Figure 1). By analyzing neural activations, we can (A) identify layers and inputs that maximally represent assertiveness, (B) examine how different sub-components interact, and (C) use steering vectors to test their causal influence on model behavior.

Prior theoretical work \citep{elhage2022superposition} suggests that related features in neural networks can be encoded as correlated but orthogonal internal representations. For instance, two concepts, such as certainty and persuasion, may often co-occur in language but be stored in distinct, minimally overlapping directions in the model’s activation space. This separation allows the network to adjust one without disrupting the other. Yet, this phenomenon has not been systematically tested for socially meaningful traits like assertiveness.

\begin{figure}
  \centering
  \includegraphics[width=0.9\linewidth]{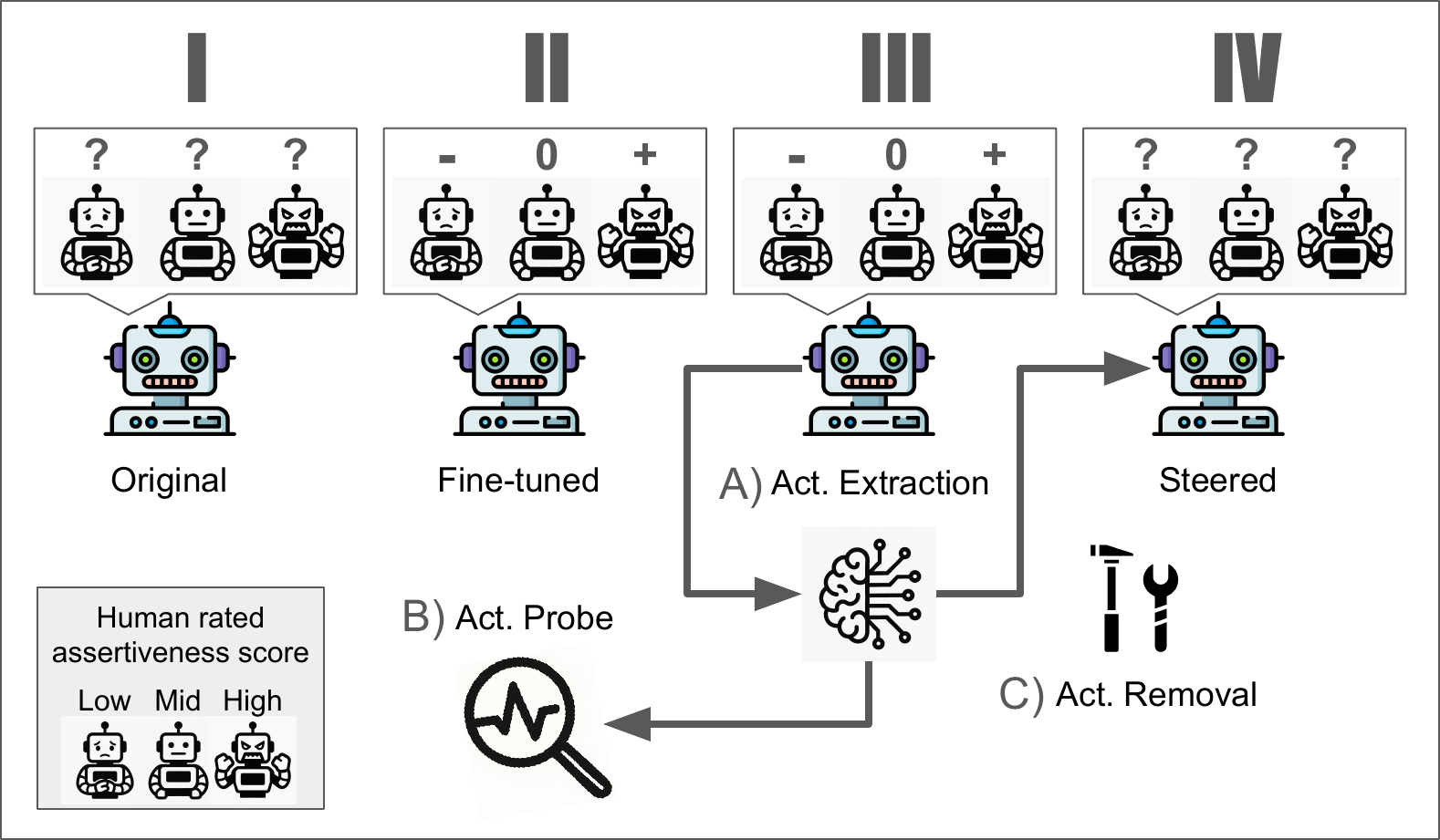}
  \caption{
  Overview of study design. Phase I: load the base LLM. Phase II: fine-tune on human rated assertiveness scores. Phase III: extract residual activations and cluster samples by similarity to identify assertiveness sub-components (A–B). Phase IV: derive and remove steering vectors for each sub-component and evaluate effects on assertiveness prediction (C). \protect\footnotemark
  }
  \label{fig:study-design}
\end{figure}

\footnotetext{
Icon credit: Colored bot icon created by Freepik — Flaticon (\url{https://www.flaticon.com/free-icon/robot_3398567}))
}

To test these claims, we hypothesize that LLMs represent assertiveness through orthogonal sub-components, notably certainty and persuasion. To evaluate this, we adopt a two-step approach: (1) clustering text samples by activation similarity to uncover latent feature categories (unsupervised), and (2) manipulating their corresponding steering vectors to assess causal effects on model predictions. This approach provides a mechanistic view of how LLMs represent linguistic assertiveness and whether its underlying components can be independently controlled.

\section{Methods}

\subsection{Datasets and model fine-tuning (Phase I-II)}

This study used the publicly available assertiveness datasets (n $=$ 645) from \citet{ghafouri_epistemic_nodate}, which were sourced from Anthropic's Persuasion dataset \citep{durmus_measuring_2024}, Globe and Mail (GM) Comments \citep{kolhatkar_sfu_2020}, Reddit Change My View (CMV) \citep{wiegmann_webis-persuasive-debaters--reddit-cmv-2022}, Llama 3–8B-generated arguments on LlAR data \citep{grattafiori_llama_2024} and Pei Assertiveness dataset \citep{pei_measuring_2021}. These datasets contained standardized assertiveness scores rated by expert annotators, with strong inter-annotator agreement (r $>$ 0.7) verified by secondary raters.

\begin{figure}
  \centering
  \includegraphics[width=0.85\linewidth]{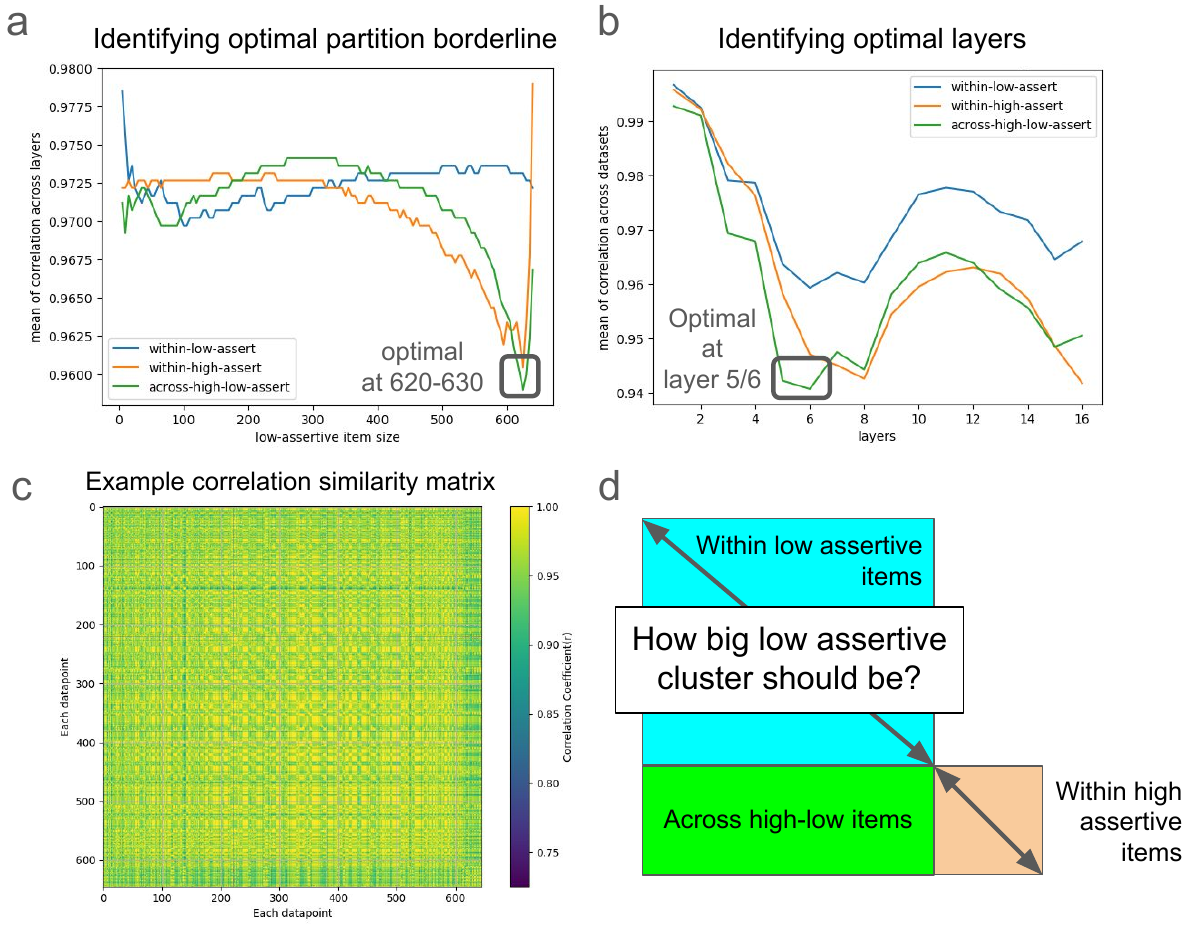}
  \caption{
  (a) Empirical identification of an optimal partition borderline between low- and high-assertive items; the green curve minimum indicates maximal separation. (b) Layer-wise analysis; the green curve minimum marks the layer with strongest low–high separation. (c) Example correlation similarity matrix across assertive items; brighter values indicate higher correlation. (d) Simplified clusters of (c): within low-assertive items (cyan), across high–low items (green), and within high-assertive items (orange). Each cluster computes the mean correlation of its items.}
  \label{fig:similarity-matrix}
\end{figure}

We fine-tuned a Llama-3.2-1B-Instruct model \citep{grattafiori_llama_2024} with Low-Rank Adaptation (LoRA \citep{hu_lora_2021}) to predict the assertiveness scores. Datasets were separated into training and validation sets in each fold, one entire dataset (e.g. GM) served as validation ($\approx$ 130 items) and the remaining as training ($\approx$ 515 items) for five folds. Model performance was reported as mean squared error (MSE). After 30 epochs, the mean minimal MSE across the five folds reached 1.94, reducing from 61.11 at the first epoch (See Appendix A.1.1 for details).

\subsection{Activation preprocessing and mechanistic interpretability (Phase III-IV)}

We fed all samples through the fine-tuned model and extracted per-token activations across all 16 transformer layers, focusing on residual streams at the start of each layer. To reduce dataset-source bias, we removed source-related steering vectors using the Difference-of-Means method \citep{turner_steering_2024, arditi_refusal_2024}.

Correlation similarity matrices were then computed across samples, ordered by standardized assertiveness scores (Fig. 2c) and highlighting three key regions: (i) within-low and (ii) within-high assertive clusters and (iii) across high–low pairs (Fig. 2d). By varying the partition borderline between (i) and (ii), we identified the point that minimized cross-cluster similarity, indicating maximal representational separation between low and high assertiveness (Fig. 2a). The layer where this separation peaked was taken as the most sensitive to assertiveness (Fig. 2b).

To probe internal structure, we projected activations of the high-assertive items into a t-SNE map. Spatial groupings were examined to reveal data-driven sub-components (e.g., certainty vs persuasion) for later conceptual interpretation. Steering vectors from each sub-component were selectively removed, and Root Mean Squared Error (RMSE) changes in assertive prediction were measured to test whether these representations function independently or interact (see Appendix A.1.2 for details).

\section{Results}

Our method identified an optimal borderline separating low- (n $=$ 620) and high-assertive items (n $=$ 25) (Fig. 2a), with middle layers (5–6) exhibiting maximal representational separation (Fig. 2b).

With this borderline and layer setting, t-SNE visualizations revealed two distinct clusters within the high-assertive items (n $=$ 25) (Figure 3a). Based on ChatGPT summaries, one cluster was labeled “emotional (big green circle)” and the other “logical (big purple circle),” aligning with the Elaboration Likelihood Model, a long-standing dual persuasion model in Psychology \citep{petty_personal_nodate_1981}. The theory posits that the logical cluster corresponds to central-route persuasion (evidence, statistics, facts), while the emotional cluster corresponds to peripheral-route persuasion (affective or superficial cues). The logical sub-component partially overlapped with certainty expressions, but which had minor influence on model predictions in these datasets.

Adding extra 25 low-assertive items (blue indices) in the t-SNE map showed partial overlap with the emotional cluster but not the logical one (Figure 3a), suggesting partial independence of the logical sub-component. High-assertive sub-components were spatially cohesive and clearly separating from low-assertive items, reflecting a decision boundary in assertiveness representation space.

Removing the emotional vector broadly degraded performance statistically (i.e. changed beyond SEM error bars), particularly on all low-assertive items and emotionally-relevant high-assertive cluster. In contrast, removing the logical vector primarily affected predictions for logical high-assertive items, indicating a more localized effect (Figure 3b).

\begin{figure}
  \centering
  \includegraphics[width=0.9\linewidth]{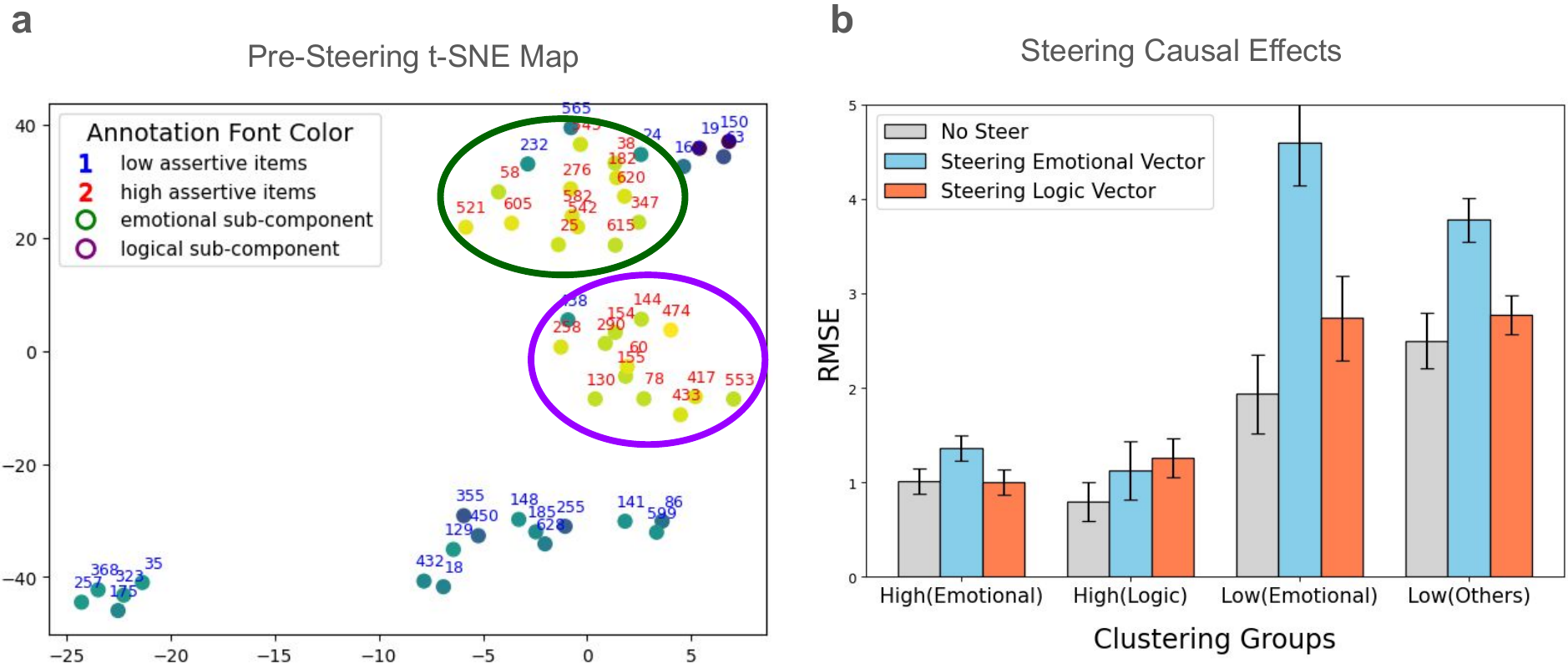}
  \caption{(a) t-SNE visualization of assertiveness clusters, showing low-assertive items (blue indices) and high-assertive items (red indices). Green and purple circles highlight emotional and logical sub-components, respectively (manual labels based on local grouping). Circle fill color indicates standardized assertive score, with lighter shades for higher scores. (b) RMSE changes after removing emotional or logical sub-component vectors, shown separately for high-assertive-logical, high-assertive-emotional, low-assertive-emotional (e.g. index 232, 16, 19), and other low-assertive items (e.g. index 35, 432, 141). Error bars are standard error of the mean (SEM).}
  \label{fig:tsne-results}
\end{figure}

\section{Conclusion and limitations}

We present an empirical method for localizing and interpreting LLM activations linked to assertiveness along a continuous scale. Results show that high-assertive representations split into two orthogonal sub-components—emotional and logical persuasion—each exerting distinct effects when used as steering vectors. Yet, this study has several limitations. The model analyzed is relatively small (1B parameters), leaving scaling effects uncertain, and only one architecture is examined, so cross-model generality remains untested. The dataset size is modest (n = 645), which may obscure finer-grained cues. Moreover, cluster interpretation relies on manual inspection, suggesting that future work should develop more automated approaches. Despite these limitations, this study provides the first visualization of correlated feature decomposition along a continuous assertiveness scale. It offers new methodological insights for analyzing LLM internal representations and contributes to understanding the roots of overconfident behaviors in LLMs.

\clearpage

\section{Acknowledgements}

This work was partially completed during the SPAR (Supervised Program for Alignment Research) programme. We are especially grateful to Kellin Pelrine (Mila/FAR.AI), Dr. Maximilian Puelma Touzel (Mila/University of Montreal), Prof. Jean-François Godbout (Mila/University of Montreal), and Bijean Ghafouri (University of Southern California) for their mentorship, guidance, and valuable feedback throughout this project.

\clearpage


\begin{thebibliography}{21}
\providecommand{\natexlab}[1]{#1}
\providecommand{\url}[1]{\texttt{#1}}
\expandafter\ifx\csname urlstyle\endcsname\relax
  \providecommand{\doi}[1]{doi: #1}\else
  \providecommand{\doi}{doi: \begingroup \urlstyle{rm}\Url}\fi

\bibitem[Arditi et~al.(2024)Arditi, Obeso, Syed, Paleka, Panickssery, Gurnee,
  and Nanda]{arditi_refusal_2024}
Andy Arditi, Oscar Obeso, Aaquib Syed, Daniel Paleka, Nina Panickssery, Wes
  Gurnee, and Neel Nanda.
\newblock Refusal in language models is mediated by a single direction, 2024.
\newblock URL \url{http://arxiv.org/abs/2406.11717}.

\bibitem[Bai et~al.(2025)Bai, Voelkel, Muldowney, Eichstaedt, and
  Willer]{bai_llm-generated_2025}
Hui Bai, Jan~G. Voelkel, Shane Muldowney, Johannes~C. Eichstaedt, and Robb
  Willer.
\newblock {LLM}-generated messages can persuade humans on policy issues.
\newblock 16\penalty0 (1):\penalty0 6037, 2025.
\newblock ISSN 2041-1723.
\newblock \doi{10.1038/s41467-025-61345-5}.
\newblock URL \url{https://www.nature.com/articles/s41467-025-61345-5}.

\bibitem[Durmus et~al.(2024)Durmus, Lovitt, Tamkin, Ritchie, Clark, and
  Ganguli]{durmus_measuring_2024}
Esin Durmus, Liane Lovitt, Alex Tamkin, Stuart Ritchie, Jack Clark, and Deep
  Ganguli.
\newblock Measuring the persuasiveness of language models, 2024.
\newblock URL
  \url{https://www.anthropic.com/news/measuring-model-persuasiveness}.

\bibitem[Elhage et~al.(2022)Elhage, Hume, Olsson, Schiefer, Henighan, Kravec,
  Hatfield-Dodds, Lasenby, Drain, Chen, Grosse, McCandlish, Kaplan, Amodei,
  Wattenberg, and Olah]{elhage2022superposition}
Nelson Elhage, Tristan Hume, Catherine Olsson, Nicholas Schiefer, Tom Henighan,
  Shauna Kravec, Zac Hatfield-Dodds, Robert Lasenby, Dawn Drain, Carol Chen,
  Roger Grosse, Sam McCandlish, Jared Kaplan, Dario Amodei, Martin Wattenberg,
  and Christopher Olah.
\newblock Toy models of superposition.
\newblock \emph{Transformer Circuits Thread}, 2022.
\newblock https://transformer-circuits.pub/2022/toy\_model/index.html.

\bibitem[Ghafouri et~al.(2024)Ghafouri, Mohammadzadeh, Zhou, Nair, Tian,
  Tsujimura, Goel, Krishna, Rabbany, Godbout, and
  Pelrine]{ghafouri_epistemic_nodate}
Bijean Ghafouri, Shahrad Mohammadzadeh, James Zhou, Pratheeksha Nair,
  Jacob-Junqi Tian, Hikaru Tsujimura, Mayank Goel, Sukanya Krishna, Reihaneh
  Rabbany, Jean-François Godbout, and Kellin Pelrine.
\newblock {EPISTEMIC} {INTEGRITY} {IN} {LARGE} {LANGUAGE} {MODELS}.
\newblock 2024.

\bibitem[Grattafiori et~al.(2024)Grattafiori, Dubey, Jauhri, Pandey, Kadian,
  Al-Dahle, Letman, Mathur, Schelten, Vaughan, Yang, Fan, Goyal, Hartshorn,
  Yang, Mitra, Sravankumar, Korenev, Hinsvark, Rao, Zhang, Rodriguez,
  Gregerson, Spataru, Roziere, Biron, Tang, Chern, Caucheteux, Nayak, Bi,
  Marra, {McConnell}, Keller, Touret, Wu, Wong, Ferrer, Nikolaidis, Allonsius,
  Song, Pintz, Livshits, Wyatt, Esiobu, Choudhary, Mahajan, Garcia-Olano,
  Perino, Hupkes, Lakomkin, {AlBadawy}, Lobanova, Dinan, Smith, Radenovic,
  Guzmán, Zhang, Synnaeve, Lee, Anderson, Thattai, Nail, Mialon, Pang,
  Cucurell, Nguyen, Korevaar, Xu, Touvron, Zarov, Ibarra, Kloumann, Misra,
  Evtimov, Zhang, Copet, Lee, Geffert, Vranes, Park, Mahadeokar, Shah, Linde,
  Billock, Hong, Lee, Fu, Chi, Huang, Liu, Wang, Yu, Bitton, Spisak, Park,
  Rocca, Johnstun, Saxe, Jia, Alwala, Prasad, Upasani, Plawiak, Li, Heafield,
  Stone, El-Arini, Iyer, Malik, Chiu, Bhalla, Lakhotia, Rantala-Yeary, Maaten,
  Chen, Tan, Jenkins, Martin, Madaan, Malo, Blecher, Landzaat, Oliveira, Muzzi,
  Pasupuleti, Singh, Paluri, Kardas, Tsimpoukelli, Oldham, Rita, Pavlova,
  Kambadur, Lewis, Si, Singh, Hassan, Goyal, Torabi, Bashlykov, Bogoychev,
  Chatterji, Zhang, Duchenne, Çelebi, Alrassy, Zhang, Li, Vasic, Weng,
  Bhargava, Dubal, Krishnan, Koura, Xu, He, Dong, Srinivasan, Ganapathy,
  Calderer, Cabral, Stojnic, Raileanu, Maheswari, Girdhar, Patel, Sauvestre,
  Polidoro, Sumbaly, Taylor, Silva, Hou, Wang, Hosseini, Chennabasappa, Singh,
  Bell, Kim, Edunov, Nie, Narang, Raparthy, Shen, Wan, Bhosale, Zhang,
  Vandenhende, Batra, Whitman, Sootla, Collot, Gururangan, Borodinsky, Herman,
  Fowler, Sheasha, Georgiou, Scialom, Speckbacher, Mihaylov, Xiao, Karn,
  Goswami, Gupta, Ramanathan, Kerkez, Gonguet, Do, Vogeti, Albiero, Petrovic,
  Chu, Xiong, Fu, Meers, Martinet, Wang, Wang, Tan, Xia, Xie, Jia, Wang,
  Goldschlag, Gaur, Babaei, Wen, Song, Zhang, Li, Mao, Coudert, Yan, Chen,
  Papakipos, Singh, Srivastava, Jain, Kelsey, Shajnfeld, Gangidi, Victoria,
  Goldstand, Menon, Sharma, Boesenberg, Baevski, Feinstein, Kallet, Sangani,
  Teo, Yunus, Lupu, Alvarado, Caples, Gu, Ho, Poulton, Ryan, Ramchandani, Dong,
  Franco, Goyal, Saraf, Chowdhury, Gabriel, Bharambe, Eisenman, Yazdan, James,
  Maurer, Leonhardi, Huang, Loyd, Paola, Paranjape, Liu, Wu, Ni, Hancock,
  Wasti, Spence, Stojkovic, Gamido, Montalvo, Parker, Burton, Mejia, Liu, Wang,
  Kim, Zhou, Hu, Chu, Cai, Tindal, Feichtenhofer, Gao, Civin, Beaty, Kreymer,
  Li, Adkins, Xu, Testuggine, David, Parikh, Liskovich, Foss, Wang, Le,
  Holland, Dowling, Jamil, Montgomery, Presani, Hahn, Wood, Le, Brinkman,
  Arcaute, Dunbar, Smothers, Sun, Kreuk, Tian, Kokkinos, Ozgenel, Caggioni,
  Kanayet, Seide, Florez, Schwarz, Badeer, Swee, Halpern, Herman, Sizov,
  Guangyi, Zhang, Lakshminarayanan, Inan, Shojanazeri, Zou, Wang, Zha, Habeeb,
  Rudolph, Suk, Aspegren, Goldman, Zhan, Damlaj, Molybog, Tufanov, Leontiadis,
  Veliche, Gat, Weissman, Geboski, Kohli, Lam, Asher, Gaya, Marcus, Tang, Chan,
  Zhen, Reizenstein, Teboul, Zhong, Jin, Yang, Cummings, Carvill, Shepard,
  {McPhie}, Torres, Ginsburg, Wang, Wu, U, Saxena, Khandelwal, Zand, Matosich,
  Veeraraghavan, Michelena, Li, Jagadeesh, Huang, Chawla, Huang, Chen, Garg, A,
  Silva, Bell, Zhang, Guo, Yu, Moshkovich, Wehrstedt, Khabsa, Avalani, Bhatt,
  Mankus, Hasson, Lennie, Reso, Groshev, Naumov, Lathi, Keneally, Liu, Seltzer,
  Valko, Restrepo, Patel, Vyatskov, Samvelyan, Clark, Macey, Wang, Hermoso,
  Metanat, Rastegari, Bansal, Santhanam, Parks, White, Bawa, Singhal, Egebo,
  Usunier, Mehta, Laptev, Dong, Cheng, Chernoguz, Hart, Salpekar, Kalinli,
  Kent, Parekh, Saab, Balaji, Rittner, Bontrager, Roux, Dollar, Zvyagina,
  Ratanchandani, Yuvraj, Liang, Alao, Rodriguez, Ayub, Murthy, Nayani, Mitra,
  Parthasarathy, Li, Hogan, Battey, Wang, Howes, Rinott, Mehta, Siby, Bondu,
  Datta, Chugh, Hunt, Dhillon, Sidorov, Pan, Mahajan, Verma, Yamamoto,
  Ramaswamy, Lindsay, Lindsay, Feng, Lin, Zha, Patil, Shankar, Zhang, Zhang,
  Wang, Agarwal, Sajuyigbe, Chintala, Max, Chen, Kehoe, Satterfield,
  Govindaprasad, Gupta, Deng, Cho, Virk, Subramanian, Choudhury, Goldman,
  Remez, Glaser, Best, Koehler, Robinson, Li, Zhang, Matthews, Chou, Shaked,
  Vontimitta, Ajayi, Montanez, Mohan, Kumar, Mangla, Ionescu, Poenaru,
  Mihailescu, Ivanov, Li, Wang, Jiang, Bouaziz, Constable, Tang, Wu, Wang, Wu,
  Gao, Kleinman, Chen, Hu, Jia, Qi, Li, Zhang, Zhang, Adi, Nam, Yu, Wang, Zhao,
  Hao, Qian, Li, He, Rait, {DeVito}, Rosnbrick, Wen, Yang, Zhao, and
  Ma]{grattafiori_llama_2024}
Aaron Grattafiori, Abhimanyu Dubey, Abhinav Jauhri, Abhinav Pandey, Abhishek
  Kadian, Ahmad Al-Dahle, Aiesha Letman, Akhil Mathur, Alan Schelten, Alex
  Vaughan, Amy Yang, Angela Fan, Anirudh Goyal, Anthony Hartshorn, Aobo Yang,
  Archi Mitra, Archie Sravankumar, Artem Korenev, Arthur Hinsvark, Arun Rao,
  Aston Zhang, Aurelien Rodriguez, Austen Gregerson, Ava Spataru, Baptiste
  Roziere, Bethany Biron, Binh Tang, Bobbie Chern, Charlotte Caucheteux, Chaya
  Nayak, Chloe Bi, Chris Marra, Chris {McConnell}, Christian Keller, Christophe
  Touret, Chunyang Wu, Corinne Wong, Cristian~Canton Ferrer, Cyrus Nikolaidis,
  Damien Allonsius, Daniel Song, Danielle Pintz, Danny Livshits, Danny Wyatt,
  David Esiobu, Dhruv Choudhary, Dhruv Mahajan, Diego Garcia-Olano, Diego
  Perino, Dieuwke Hupkes, Egor Lakomkin, Ehab {AlBadawy}, Elina Lobanova, Emily
  Dinan, Eric~Michael Smith, Filip Radenovic, Francisco Guzmán, Frank Zhang,
  Gabriel Synnaeve, Gabrielle Lee, Georgia~Lewis Anderson, Govind Thattai,
  Graeme Nail, Gregoire Mialon, Guan Pang, Guillem Cucurell, Hailey Nguyen,
  Hannah Korevaar, Hu~Xu, Hugo Touvron, Iliyan Zarov, Imanol~Arrieta Ibarra,
  Isabel Kloumann, Ishan Misra, Ivan Evtimov, Jack Zhang, Jade Copet, Jaewon
  Lee, Jan Geffert, Jana Vranes, Jason Park, Jay Mahadeokar, Jeet Shah, Jelmer
  van~der Linde, Jennifer Billock, Jenny Hong, Jenya Lee, Jeremy Fu, Jianfeng
  Chi, Jianyu Huang, Jiawen Liu, Jie Wang, Jiecao Yu, Joanna Bitton, Joe
  Spisak, Jongsoo Park, Joseph Rocca, Joshua Johnstun, Joshua Saxe, Junteng
  Jia, Kalyan~Vasuden Alwala, Karthik Prasad, Kartikeya Upasani, Kate Plawiak,
  Ke~Li, Kenneth Heafield, Kevin Stone, Khalid El-Arini, Krithika Iyer, Kshitiz
  Malik, Kuenley Chiu, Kunal Bhalla, Kushal Lakhotia, Lauren Rantala-Yeary,
  Laurens van~der Maaten, Lawrence Chen, Liang Tan, Liz Jenkins, Louis Martin,
  Lovish Madaan, Lubo Malo, Lukas Blecher, Lukas Landzaat, Luke~de Oliveira,
  Madeline Muzzi, Mahesh Pasupuleti, Mannat Singh, Manohar Paluri, Marcin
  Kardas, Maria Tsimpoukelli, Mathew Oldham, Mathieu Rita, Maya Pavlova,
  Melanie Kambadur, Mike Lewis, Min Si, Mitesh~Kumar Singh, Mona Hassan, Naman
  Goyal, Narjes Torabi, Nikolay Bashlykov, Nikolay Bogoychev, Niladri
  Chatterji, Ning Zhang, Olivier Duchenne, Onur Çelebi, Patrick Alrassy,
  Pengchuan Zhang, Pengwei Li, Petar Vasic, Peter Weng, Prajjwal Bhargava,
  Pratik Dubal, Praveen Krishnan, Punit~Singh Koura, Puxin Xu, Qing He,
  Qingxiao Dong, Ragavan Srinivasan, Raj Ganapathy, Ramon Calderer,
  Ricardo~Silveira Cabral, Robert Stojnic, Roberta Raileanu, Rohan Maheswari,
  Rohit Girdhar, Rohit Patel, Romain Sauvestre, Ronnie Polidoro, Roshan
  Sumbaly, Ross Taylor, Ruan Silva, Rui Hou, Rui Wang, Saghar Hosseini, Sahana
  Chennabasappa, Sanjay Singh, Sean Bell, Seohyun~Sonia Kim, Sergey Edunov,
  Shaoliang Nie, Sharan Narang, Sharath Raparthy, Sheng Shen, Shengye Wan,
  Shruti Bhosale, Shun Zhang, Simon Vandenhende, Soumya Batra, Spencer Whitman,
  Sten Sootla, Stephane Collot, Suchin Gururangan, Sydney Borodinsky, Tamar
  Herman, Tara Fowler, Tarek Sheasha, Thomas Georgiou, Thomas Scialom, Tobias
  Speckbacher, Todor Mihaylov, Tong Xiao, Ujjwal Karn, Vedanuj Goswami, Vibhor
  Gupta, Vignesh Ramanathan, Viktor Kerkez, Vincent Gonguet, Virginie Do, Vish
  Vogeti, Vítor Albiero, Vladan Petrovic, Weiwei Chu, Wenhan Xiong, Wenyin Fu,
  Whitney Meers, Xavier Martinet, Xiaodong Wang, Xiaofang Wang, Xiaoqing~Ellen
  Tan, Xide Xia, Xinfeng Xie, Xuchao Jia, Xuewei Wang, Yaelle Goldschlag,
  Yashesh Gaur, Yasmine Babaei, Yi~Wen, Yiwen Song, Yuchen Zhang, Yue Li,
  Yuning Mao, Zacharie~Delpierre Coudert, Zheng Yan, Zhengxing Chen, Zoe
  Papakipos, Aaditya Singh, Aayushi Srivastava, Abha Jain, Adam Kelsey, Adam
  Shajnfeld, Adithya Gangidi, Adolfo Victoria, Ahuva Goldstand, Ajay Menon,
  Ajay Sharma, Alex Boesenberg, Alexei Baevski, Allie Feinstein, Amanda Kallet,
  Amit Sangani, Amos Teo, Anam Yunus, Andrei Lupu, Andres Alvarado, Andrew
  Caples, Andrew Gu, Andrew Ho, Andrew Poulton, Andrew Ryan, Ankit Ramchandani,
  Annie Dong, Annie Franco, Anuj Goyal, Aparajita Saraf, Arkabandhu Chowdhury,
  Ashley Gabriel, Ashwin Bharambe, Assaf Eisenman, Azadeh Yazdan, Beau James,
  Ben Maurer, Benjamin Leonhardi, Bernie Huang, Beth Loyd, Beto~De Paola,
  Bhargavi Paranjape, Bing Liu, Bo~Wu, Boyu Ni, Braden Hancock, Bram Wasti,
  Brandon Spence, Brani Stojkovic, Brian Gamido, Britt Montalvo, Carl Parker,
  Carly Burton, Catalina Mejia, Ce~Liu, Changhan Wang, Changkyu Kim, Chao Zhou,
  Chester Hu, Ching-Hsiang Chu, Chris Cai, Chris Tindal, Christoph
  Feichtenhofer, Cynthia Gao, Damon Civin, Dana Beaty, Daniel Kreymer, Daniel
  Li, David Adkins, David Xu, Davide Testuggine, Delia David, Devi Parikh,
  Diana Liskovich, Didem Foss, Dingkang Wang, Duc Le, Dustin Holland, Edward
  Dowling, Eissa Jamil, Elaine Montgomery, Eleonora Presani, Emily Hahn, Emily
  Wood, Eric-Tuan Le, Erik Brinkman, Esteban Arcaute, Evan Dunbar, Evan
  Smothers, Fei Sun, Felix Kreuk, Feng Tian, Filippos Kokkinos, Firat Ozgenel,
  Francesco Caggioni, Frank Kanayet, Frank Seide, Gabriela~Medina Florez,
  Gabriella Schwarz, Gada Badeer, Georgia Swee, Gil Halpern, Grant Herman,
  Grigory Sizov, Guangyi, Zhang, Guna Lakshminarayanan, Hakan Inan, Hamid
  Shojanazeri, Han Zou, Hannah Wang, Hanwen Zha, Haroun Habeeb, Harrison
  Rudolph, Helen Suk, Henry Aspegren, Hunter Goldman, Hongyuan Zhan, Ibrahim
  Damlaj, Igor Molybog, Igor Tufanov, Ilias Leontiadis, Irina-Elena Veliche,
  Itai Gat, Jake Weissman, James Geboski, James Kohli, Janice Lam, Japhet
  Asher, Jean-Baptiste Gaya, Jeff Marcus, Jeff Tang, Jennifer Chan, Jenny Zhen,
  Jeremy Reizenstein, Jeremy Teboul, Jessica Zhong, Jian Jin, Jingyi Yang, Joe
  Cummings, Jon Carvill, Jon Shepard, Jonathan {McPhie}, Jonathan Torres, Josh
  Ginsburg, Junjie Wang, Kai Wu, Kam~Hou U, Karan Saxena, Kartikay Khandelwal,
  Katayoun Zand, Kathy Matosich, Kaushik Veeraraghavan, Kelly Michelena, Keqian
  Li, Kiran Jagadeesh, Kun Huang, Kunal Chawla, Kyle Huang, Lailin Chen,
  Lakshya Garg, Lavender A, Leandro Silva, Lee Bell, Lei Zhang, Liangpeng Guo,
  Licheng Yu, Liron Moshkovich, Luca Wehrstedt, Madian Khabsa, Manav Avalani,
  Manish Bhatt, Martynas Mankus, Matan Hasson, Matthew Lennie, Matthias Reso,
  Maxim Groshev, Maxim Naumov, Maya Lathi, Meghan Keneally, Miao Liu,
  Michael~L. Seltzer, Michal Valko, Michelle Restrepo, Mihir Patel, Mik
  Vyatskov, Mikayel Samvelyan, Mike Clark, Mike Macey, Mike Wang, Miquel~Jubert
  Hermoso, Mo~Metanat, Mohammad Rastegari, Munish Bansal, Nandhini Santhanam,
  Natascha Parks, Natasha White, Navyata Bawa, Nayan Singhal, Nick Egebo,
  Nicolas Usunier, Nikhil Mehta, Nikolay~Pavlovich Laptev, Ning Dong, Norman
  Cheng, Oleg Chernoguz, Olivia Hart, Omkar Salpekar, Ozlem Kalinli, Parkin
  Kent, Parth Parekh, Paul Saab, Pavan Balaji, Pedro Rittner, Philip Bontrager,
  Pierre Roux, Piotr Dollar, Polina Zvyagina, Prashant Ratanchandani, Pritish
  Yuvraj, Qian Liang, Rachad Alao, Rachel Rodriguez, Rafi Ayub, Raghotham
  Murthy, Raghu Nayani, Rahul Mitra, Rangaprabhu Parthasarathy, Raymond Li,
  Rebekkah Hogan, Robin Battey, Rocky Wang, Russ Howes, Ruty Rinott, Sachin
  Mehta, Sachin Siby, Sai~Jayesh Bondu, Samyak Datta, Sara Chugh, Sara Hunt,
  Sargun Dhillon, Sasha Sidorov, Satadru Pan, Saurabh Mahajan, Saurabh Verma,
  Seiji Yamamoto, Sharadh Ramaswamy, Shaun Lindsay, Shaun Lindsay, Sheng Feng,
  Shenghao Lin, Shengxin~Cindy Zha, Shishir Patil, Shiva Shankar, Shuqiang
  Zhang, Shuqiang Zhang, Sinong Wang, Sneha Agarwal, Soji Sajuyigbe, Soumith
  Chintala, Stephanie Max, Stephen Chen, Steve Kehoe, Steve Satterfield,
  Sudarshan Govindaprasad, Sumit Gupta, Summer Deng, Sungmin Cho, Sunny Virk,
  Suraj Subramanian, Sy~Choudhury, Sydney Goldman, Tal Remez, Tamar Glaser,
  Tamara Best, Thilo Koehler, Thomas Robinson, Tianhe Li, Tianjun Zhang, Tim
  Matthews, Timothy Chou, Tzook Shaked, Varun Vontimitta, Victoria Ajayi,
  Victoria Montanez, Vijai Mohan, Vinay~Satish Kumar, Vishal Mangla, Vlad
  Ionescu, Vlad Poenaru, Vlad~Tiberiu Mihailescu, Vladimir Ivanov, Wei Li,
  Wenchen Wang, Wenwen Jiang, Wes Bouaziz, Will Constable, Xiaocheng Tang,
  Xiaojian Wu, Xiaolan Wang, Xilun Wu, Xinbo Gao, Yaniv Kleinman, Yanjun Chen,
  Ye~Hu, Ye~Jia, Ye~Qi, Yenda Li, Yilin Zhang, Ying Zhang, Yossi Adi, Youngjin
  Nam, Yu, Wang, Yu~Zhao, Yuchen Hao, Yundi Qian, Yunlu Li, Yuzi He, Zach Rait,
  Zachary {DeVito}, Zef Rosnbrick, Zhaoduo Wen, Zhenyu Yang, Zhiwei Zhao, and
  Zhiyu Ma.
\newblock The llama 3 herd of models, 2024.
\newblock URL \url{http://arxiv.org/abs/2407.21783}.

\bibitem[Hu et~al.(2021)Hu, Shen, Wallis, Allen-Zhu, Li, Wang, Wang, and
  Chen]{hu_lora_2021}
Edward~J. Hu, Yelong Shen, Phillip Wallis, Zeyuan Allen-Zhu, Yuanzhi Li, Shean
  Wang, Lu~Wang, and Weizhu Chen.
\newblock {LoRA}: Low-rank adaptation of large language models, 2021.
\newblock URL \url{http://arxiv.org/abs/2106.09685}.

\bibitem[Karelitz and Budescu(2004)]{karelitz_you_2004}
Tzur~M. Karelitz and David~V. Budescu.
\newblock You say "probable" and i say "likely": Improving interpersonal
  communication with verbal probability phrases.
\newblock 10\penalty0 (1):\penalty0 25--41, 2004.
\newblock ISSN 1939-2192, 1076-898X.
\newblock \doi{10.1037/1076-898X.10.1.25}.
\newblock URL \url{https://doi.apa.org/doi/10.1037/1076-898X.10.1.25}.

\bibitem[Kolhatkar et~al.(2020)Kolhatkar, Wu, Cavasso, Francis, Shukla, and
  Taboada]{kolhatkar_sfu_2020}
Varada Kolhatkar, Hanhan Wu, Luca Cavasso, Emilie Francis, Kavan Shukla, and
  Maite Taboada.
\newblock The {SFU} opinion and comments corpus: A corpus for the analysis of
  online news comments.
\newblock 4\penalty0 (2):\penalty0 155--190, 2020.
\newblock ISSN 2509-9507, 2509-9515.
\newblock \doi{10.1007/s41701-019-00065-w}.
\newblock URL \url{http://link.springer.com/10.1007/s41701-019-00065-w}.

\bibitem[Lai et~al.(2024)Lai, Gan, Wu, Qi, and Yu]{lai_large_2024}
Jinqi Lai, Wensheng Gan, Jiayang Wu, Zhenlian Qi, and Philip~S. Yu.
\newblock Large language models in law: A survey.
\newblock 5:\penalty0 181--196, 2024.
\newblock ISSN 26666510.
\newblock \doi{10.1016/j.aiopen.2024.09.002}.
\newblock URL
  \url{https://linkinghub.elsevier.com/retrieve/pii/S2666651024000172}.

\bibitem[Leng et~al.(2025)Leng, Huang, Zhu, and Huang]{leng_taming_2025}
Jixuan Leng, Chengsong Huang, Banghua Zhu, and Jiaxin Huang.
\newblock Taming overconfidence in {LLMs}: Reward calibration in {RLHF}, 2025.
\newblock URL \url{http://arxiv.org/abs/2410.09724}.

\bibitem[Paradeda et~al.(2019)Paradeda, Ferreira, Oliveira, Martinho, and
  Paiva]{paradeda_role_2019}
Raul Paradeda, Maria~José Ferreira, Raquel Oliveira, Carlos Martinho, and Ana
  Paiva.
\newblock The role of assertiveness in a storytelling game with persuasive
  robotic non-player characters.
\newblock In \emph{Proceedings of the Annual Symposium on Computer-Human
  Interaction in Play}, pages 453--465. {ACM}, 2019.
\newblock ISBN 978-1-4503-6688-5.
\newblock \doi{10.1145/3311350.3347162}.
\newblock URL \url{https://dl.acm.org/doi/10.1145/3311350.3347162}.

\bibitem[Pei and Jurgens(2021)]{pei_measuring_2021}
Jiaxin Pei and David Jurgens.
\newblock Measuring sentence-level and aspect-level (un)certainty in science
  communications, 2021.
\newblock URL \url{http://arxiv.org/abs/2109.14776}.

\bibitem[Petty et~al.(1981)Petty, Cacioppo, and
  Goldman]{petty_personal_nodate_1981}
Richard~E Petty, John~T Cacioppo, and Rachel Goldman.
\newblock Personal involvement as a determinant of argument-based persuasion.
\newblock 1981.

\bibitem[Sharma et~al.(2023)Sharma, Rushton, Lin, Wadden, Lucas, Miner, Nguyen,
  and Althoff]{sharma_cognitive_2023}
Ashish Sharma, Kevin Rushton, Inna Lin, David Wadden, Khendra Lucas, Adam
  Miner, Theresa Nguyen, and Tim Althoff.
\newblock Cognitive reframing of negative thoughts through human-language model
  interaction.
\newblock In \emph{Proceedings of the 61st Annual Meeting of the Association
  for Computational Linguistics (Volume 1: Long Papers)}, pages 9977--10000.
  Association for Computational Linguistics, 2023.
\newblock \doi{10.18653/v1/2023.acl-long.555}.
\newblock URL \url{https://aclanthology.org/2023.acl-long.555}.

\bibitem[Trust et~al.(2023)Trust, Whalen, and Mouza]{trust_editorial_nodate}
Torrey Trust, Jeromie Whalen, and Chrystalla Mouza.
\newblock Editorial: {ChatGPT}: Challenges, opportunities, and implications for
  teacher education.
\newblock 2023.

\bibitem[Turner et~al.(2024)Turner, Thiergart, Leech, Udell, Vazquez, Mini, and
  {MacDiarmid}]{turner_steering_2024}
Alexander~Matt Turner, Lisa Thiergart, Gavin Leech, David Udell, Juan~J.
  Vazquez, Ulisse Mini, and Monte {MacDiarmid}.
\newblock Steering language models with activation engineering, 2024.
\newblock URL \url{http://arxiv.org/abs/2308.10248}.

\bibitem[Wen et~al.(2024)Wen, Xu, Han, Wolfe, Wang, and
  Howe]{wen_mitigating_nodate}
Bingbing Wen, Chenjun Xu, Bin Han, Robert Wolfe, Lucy~Lu Wang, and Bill Howe.
\newblock Mitigating overconfidence in large language models: A behavioral lens
  on confidence estimation and calibration.
\newblock 2024.

\bibitem[Wiegmann et~al.(2022)Wiegmann, Al-Khatib, {Vishal Khanna}, and
  Stein]{wiegmann_webis-persuasive-debaters--reddit-cmv-2022}
Matti Wiegmann, Khalid Al-Khatib, {Vishal Khanna}, and Benno Stein.
\newblock Webis-persuasive-debaters-on-reddit-{CMV}-2022, 2022.
\newblock URL \url{https://zenodo.org/record/7034173}.

\bibitem[Windschitl and Wells(1996)]{windschitl_measuring_1996}
Paul~D. Windschitl and Gary~L. Wells.
\newblock Measuring psychological uncertainty: Verbal versus numeric methods.
\newblock 2\penalty0 (4):\penalty0 343--364, 1996.
\newblock ISSN 1939-2192, 1076-898X.
\newblock \doi{10.1037/1076-898X.2.4.343}.
\newblock URL \url{https://doi.apa.org/doi/10.1037/1076-898X.2.4.343}.

\bibitem[Zhou et~al.(2025)Zhou, Jin, Shi, and Li]{zhou_steerconf_2025}
Ziang Zhou, Tianyuan Jin, Jieming Shi, and Qing Li.
\newblock {SteerConf}: Steering {LLMs} for confidence elicitation, 2025.
\newblock URL \url{http://arxiv.org/abs/2503.02863}.

\end{thebibliography}

\end{document}